\documentclass[lettersize,journal]{IEEEtran}
\usepackage{amsmath,amsfonts}
\usepackage{algorithm}
\usepackage{array}
\usepackage[caption=false,font=footnotesize,labelfont=rm,textfont=rm]{subfig}
\usepackage{textcomp}
\usepackage{stfloats}
\usepackage{url}
\usepackage{verbatim}
\usepackage{graphicx}
\usepackage{cite}
\usepackage{algpseudocode}
\usepackage[dvipsnames]{xcolor}
\usepackage{multirow}
\usepackage{makecell}
\usepackage{amsfonts}
\usepackage{booktabs}
\usepackage{bbding}
\usepackage{xspace}
\usepackage{microtype}
\usepackage{enumitem}
\usepackage{pifont}
\usepackage{amsmath}
\usepackage{amssymb}
\usepackage{placeins}

\newcommand{\name}[0]{S-FGRM\xspace}

\makeatletter
\DeclareRobustCommand\onedot{\futurelet\@let@token\@onedot}
\def\@onedot{\ifx\@let@token.\else.\null\fi\xspace}

\def\eg{\emph{e.g}\onedot} 
\def\ie{\emph{i.e}\onedot} 
 
\def\etc{\emph{etc}\onedot} 
\def\wrt{w.r.t\onedot} 
\def\etal{\emph{et al}\onedot}

\makeatother

\usepackage{scalerel}
\usepackage{tikz}
\usetikzlibrary{svg.path}
\definecolor{orcidlogocol}{HTML}{A6CE39}
\tikzset{
    orcidlogo/.pic={
        \fill[orcidlogocol] svg{M256,128c0,70.7-57.3,128-128,128C57.3,256,0,198.7,0,128C0,57.3,57.3,0,128,0C198.7,0,256,57.3,256,128z};
        \fill[white] svg{M86.3,186.2H70.9V79.1h15.4v48.4V186.2z}
        svg{M108.9,79.1h41.6c39.6,0,57,28.3,57,53.6c0,27.5-21.5,53.6-56.8,53.6h-41.8V79.1z M124.3,172.4h24.5c34.9,0,42.9-26.5,42.9-39.7c0-21.5-13.7-39.7-43.7-39.7h-23.7V172.4z}
        svg{M88.7,56.8c0,5.5-4.5,10.1-10.1,10.1c-5.6,0-10.1-4.6-10.1-10.1c0-5.6,4.5-10.1,10.1-10.1C84.2,46.7,88.7,51.3,88.7,56.8z};
    }
}
\newcommand\orcidicon[1]{\href{https://orcid.org/#1}{\mbox{\scalerel*{
                \begin{tikzpicture}[yscale=-1,transform shape]
                \pic{orcidlogo};
                \end{tikzpicture}
            }{|}}}}
\usepackage[implicit=false]{hyperref}
\hypersetup{hidelinks,
	colorlinks=true,
	allcolors=black,
	pdfstartview=Fit,
	breaklinks=true}

\begin{document}

\title{Sampling-based Fast Gradient Rescaling Method for Highly Transferable Adversarial Attacks}

\author{
Xu Han$^1$, 
\and
Anmin Liu$^1$\IEEEauthorrefmark{1}, 
\and
Chenxuan Yao$^1$, \and
Yanbo Fan$^2$, \and 
Kun He$^1$\IEEEauthorrefmark{2} \textsuperscript{\orcidicon{0000-0001-7627-4604}}, \IEEEmembership{Senior~Member,~IEEE} 

\IEEEcompsocitemizethanks{\IEEEcompsocthanksitem Xu Han, Anmin Liu, Chenxuan Yao and Kun He are with School of Computer Science and Technology, Huazhong University of Science and Technology; 
and Hopcroft Center on Computing Science, Huazhong University of Science and Technology, Wuhan 430074,  China. The first two authors contribute equally. Corresponding author: Kun He. E-mail:brooklet60@hust.edu.cn. 
\IEEEcompsocthanksitem Yanbo Fan is with Tencent AI Lab, Shenzhen, China. 
}

\thanks{Manuscript received June 13th, 2023}
\thanks{This work was supported by National Natural Science Foundation (62076105,U22B2017).}
}


\markboth{}{Han \MakeLowercase{\textit{et al.}}: S-FGRM for Highly Transferable Adversarial Attacks } 


\maketitle

\begin{abstract}
Deep neural networks are known to be vulnerable to adversarial examples crafted by adding human-imperceptible perturbations to the benign input. After achieving nearly 100\% attack success rates in white-box setting, more focus is shifted to black-box attacks, of which the transferability of adversarial examples has gained significant attention. 
In either case, the common gradient-based methods generally use the \emph{sign} function to generate perturbations on the gradient update, that offers a roughly correct direction and has gained great success. But little work pays attention to its possible limitation. 
In this work, we observe that the deviation between the original gradient and the generated noise may lead to inaccurate gradient update estimation and suboptimal solutions for adversarial transferability. To this end, we propose a Sampling-based Fast Gradient Rescaling Method (S-FGRM). 
Specifically, we use data rescaling to substitute the \emph{sign} function without extra computational cost. We further propose a Depth First Sampling method to eliminate the fluctuation of rescaling and stabilize the gradient update. Our method could be used in any gradient-based attacks and is extensible to be integrated with various input transformation or ensemble methods to further improve the adversarial transferability. Extensive experiments on the standard ImageNet dataset show that our method could significantly boost the transferability of gradient-based attacks and outperform the state-of-the-art baselines.
\end{abstract}


\section{Introduction}
\label{sec:intro}
\IEEEPARstart{A}{long} with their incredible success in computer vision tasks, the robustness of deep neural networks (DNNs) has also raised serious concern. DNNs have been found to be vulnerable to adversarial examples~\cite{goodfellow2014explaining,szegedy2014intriguing} crafted by adding  human-imperceptible perturbations to the benign input. Worse still, adversarial examples have been demonstrated to be transferable~\cite{moosavi2017universal,papernot2017practical,liu2017delving}, indicating that adversarial examples crafted for current target model can be used to fool other black-box models, allowing for real-world black-box attacks~\cite{xu2020adversarial,sharif2016accessorize}.



\begin{figure}[!t]
    \centering
    \subfloat[Gradient directions]{\includegraphics[width=0.23\textwidth]{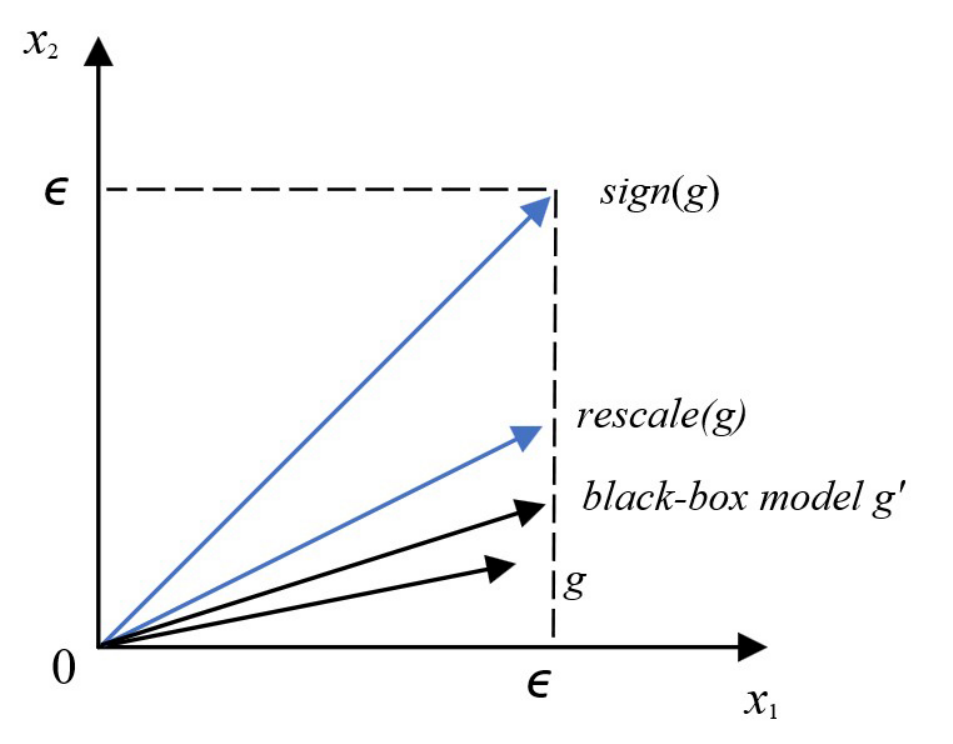}%
    \label{fig:direction}}
    \hfil
    \subfloat[Optimization paths]{\includegraphics[width=0.23\textwidth]{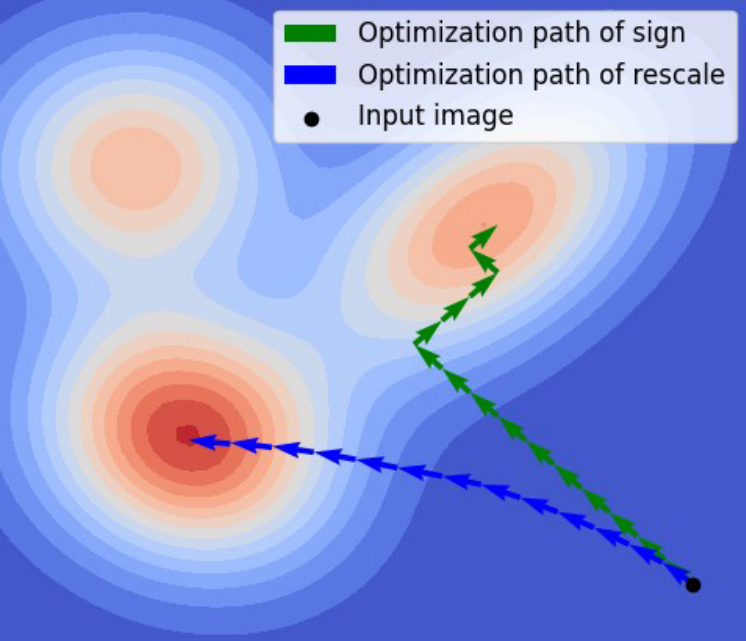}%
    \label{fig:path}}
    \caption{Precision comparison of $sign(g)$ and $rescale(g)$.}
    \label{fig:sign}
\end{figure}

Numerous works have been proposed to investigate the adversarial robustness of deep learning models, including gradient-based single-step methods such as fast gradient sign method (FGSM)~\cite{goodfellow2014explaining} and its randomized variant~\cite{tramer2018ensemble}, multi-step methods such as the basic iterative method (BIM)~\cite{kurakin2017adversarial} and the projected gradient descent (PGD) attack~\cite{madry2018towards}. These adversarial attack 
methods can achieve high success rates in white-box setting, in which the attacker has access to the architecture, model parameters and other information of the target model. However, they often exhibit low attack success rates in black-box setting for transfer attacks.

To address the above challenge, 
numerous efforts have been devoted to improve the transferability of adversarial examples, including gradient optimization attacks~\cite{kurakin2017adversarial,dong2018boosting,lin2019nesterov,wu2020ACML}, input transformation attacks~\cite{xie2019improving,dong2019evading,lin2019nesterov} and model ensemble attacks~\cite{dong2018boosting,liu2017delving,SVRE2022CVPR,hang2020ensemble}. Gradient optimization attacks attempt to boost the black-box performance by advanced gradient calculation, while input transformation attacks aim to find adversarial perturbations with higher transferability by applying various transformations to the inputs. By fusing the logit outputs, model ensemble attacks generate adversarial examples on multiple models simultaneously. The latter two categories are both based on existing gradient-based attacks to further boost the transferability.

Among these schemes, little attention has been paid to the possible limitation of the widely used basic gradient update using the \emph{sign} function. The \emph{sign} function guarantees an $l_\infty$ bound on the gradient update, as well as sufficient perturbations and a generally proper update direction, making it successful and being widely adopted in most gradient optimization methods. 
We observe that, however, the \emph{sign} function could not offer an accurate approximation on the gradient update direction, resulting in just two values of $\pm 1$ in most cases (the value of 0 rarely happens for the real value of gradients). For instance, if two dimensions of the actual gradient are $(0.8, 10^{-8})$,  the sign gradient will be $(1,1)$ but $(1,0)$ is a more accurate approximation. 
For another example, as illustrated in Fig.~\ref{fig:sign} (a) two dimensions of the gradient, where $\epsilon$ is the perturbation budget for the current update, the update direction provided by \emph{sign} deviates from the direction of both the original gradient $g$ and the actual gradient $g'$ of the black-box model more than \emph{rescale}. We argue that such inaccuracy of gradient might mislead the optimization path as shown in Fig.~\ref{fig:sign} (b) (the color shade represents the values of the function). The iterative process might fall into suboptimum compared to \emph{rescale} and weaken the adversarial transferability.

\begin{figure}[!t]
    \centering
    \subfloat[Original gradient distribution]{\includegraphics[width=0.23\textwidth]{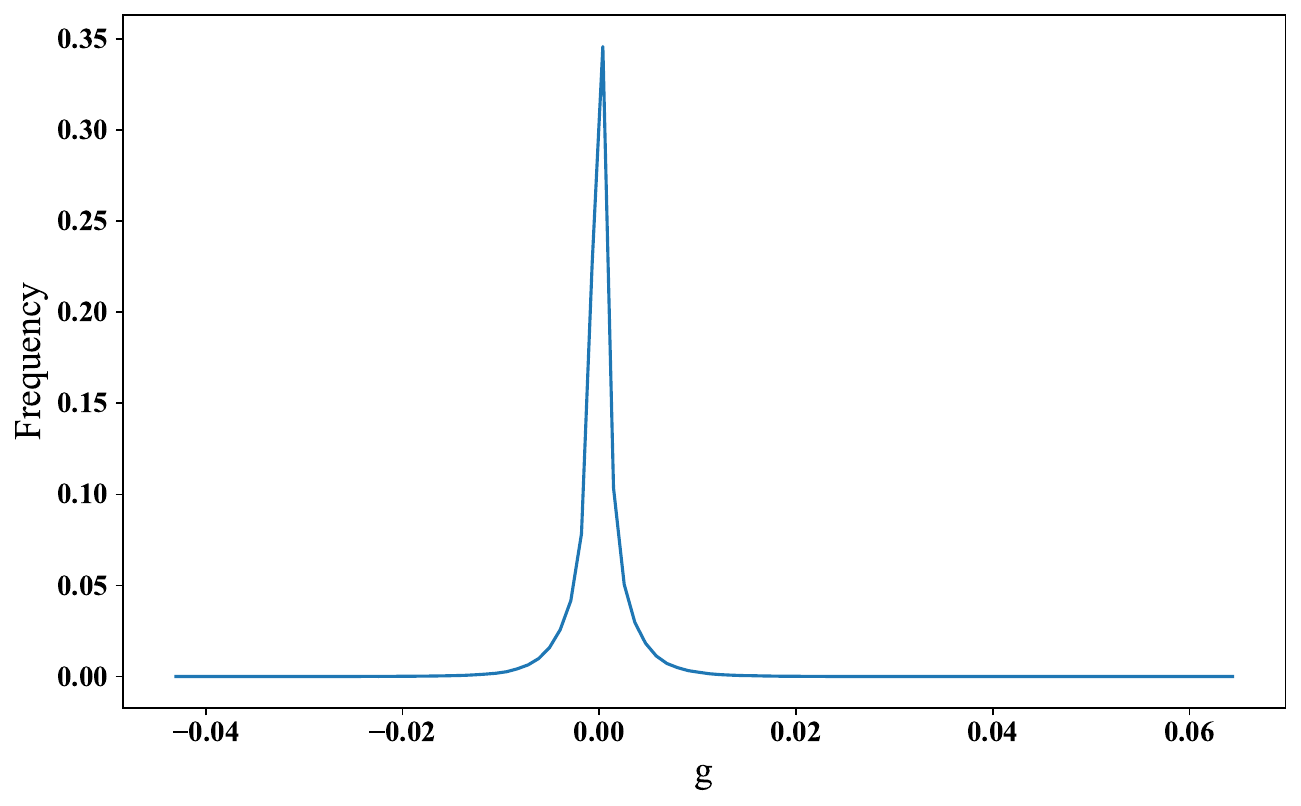}%
    \label{fig:origindistr}}
    \hfil
    \subfloat[Distribution after {\em rescale}]{\includegraphics[width=0.23\textwidth]{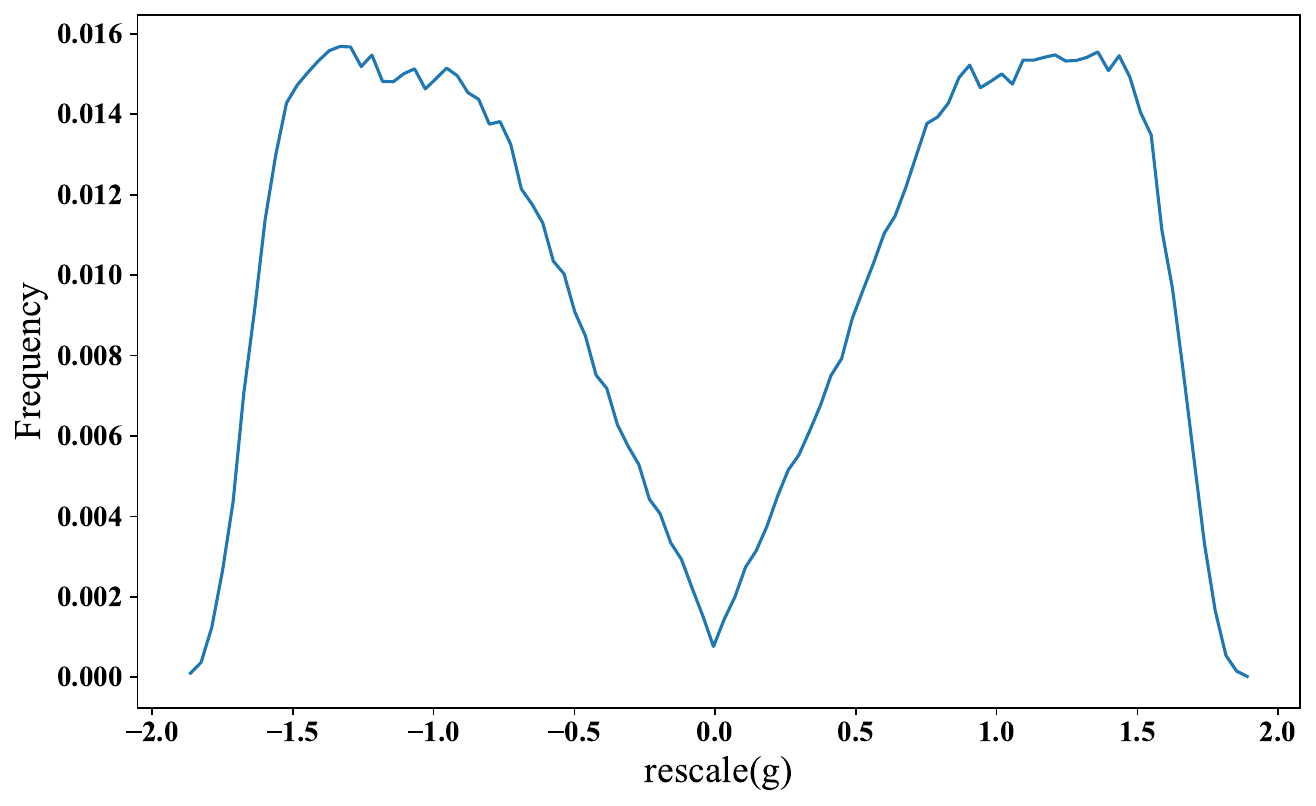}%
    \label{fig:fgrmdistr}}
    \caption{Gradient distributions before and after {\em rescale}(when $c=2$ in formula). There does exist non-negligible deviation between $sign(g)$ and the true gradient $g$. Our $rescale(g)$ could well scatter the original gradients, thus having a better chance of approaching both $g$ and $g'$ in Fig. 1.}
    \label{fig:rescaleanalysis}
    \vspace{-1pt}
\end{figure}

Based on the above analysis, we propose a Sampling-based Fast Gradient Rescaling Method (S-FGRM) to have a more accurate gradient approximation and boost the transferability of gradient-based attacks. 
Specifically, we design a data rescaling method for the gradient calculation, that directly exploits the distribution of gradient values in different dimensions and maintains their disparities.  
Moreover, as most gradient values are extremely small, as shown in Fig.~\ref{fig:rescaleanalysis} (a), we further introduce a sampling method, dubbed Depth First Sampling, to stabilize the update direction. 
Our sampling method eliminates the local fluctuation error
and further boosts the transferability of the crafted examples. 
Note that our proposed method is generally applicable to any gradient-based attack and can be combined with a variety of input transformation or ensemble methods to further enhance the adversarial transferability. 
Extensive experiments and analysis show that our integrated methods mitigate the overfitting issue and further strengthen various gradient-based attacks.

The rest of this paper is organized as follows. Section \ref{sec:RW} presents related adversarial attack methods, including gradient optimization attacks, input transformation attacks, and model ensemble attacks. Section \ref{sec:method} introduces our proposed \name, including typical gradient optimization attacks and our motivation, the details of our proposed two components, i.e., the fast gradient rescaling method (FGRM) and the depth first sampling method (DFSM), and the final \name algorithms that can be combined with either MI-FGSM or NI-FGSM. Section \ref{sec:exp} presents extensive  experimental comparisons, parameter study and ablation study. In the end, the paper summarizes with concluding remarks in Section \ref{sec:conclusion}.

\section{Related Works}
\label{sec:RW}
Due to the lack of model information, attacks in the black-box setting are more challenging. 
Researchers find that adversarial examples exhibit good transferability across the models~\cite{liu2017delving,papernot2017practical}, 
and a prevailing stream of black-box attacks is to transfer adversarial examples generated in substitute models into the target model. 
This section provides a brief overview of the three main categories of transfer-based attacks.

\subsubsection{Gradient Optimization Attacks}
Gradient optimization methods~\cite{goodfellow2014explaining,kurakin2017adversarial,madry2018towards,dong2018boosting,lin2019nesterov,wang2021enhancing}, which are the most critical category that also serves as the base for other attacks, focus on improving the attack transferability through advanced gradient calculation. 
FGSM~\cite{goodfellow2014explaining} is a single-step gradient-based attack to maximize the loss function \wrt to the input. I-FGSM~\cite{kurakin2017adversarial} extends FGSM by update the gradient iteratively. Momentum Iterative attack (MI)~\cite{dong2018boosting} and Nesterov Iterative attack (NI)~\cite{lin2019nesterov} 
introduce momentum to the iterative gradient attacks to boost the attack transferability. 
Smooth gradient attack~\cite{wu2020ACML} enhances the adversarial transferbility by smoothing the loss surface, such that the crafted examples are robust to Gaussian perturbation. 
Variance Tuning attack~\cite{wang2021enhancing} uses the gradient variance of the previous iteration to tune the current gradient calculation so as to have a more stable update direction. 
Different from the above methods that design advanced strategies upon FGSM, our \name focuses on how to overcome the limitation of the \emph{sign} function in FGSM.

\subsubsection{Input Transformation Attacks}

The second category of boosting the adversarial transferability is to adopt various input transformations. 
Diverse Input Method (DIM)~\cite{xie2019improving} applies random resizing and padding to the input to alleviate the overfitting issue of I-FGSM, resulting in a high attack success rate under both white-box and black-box settings. 
Scale-Invariant Method (SIM)~\cite{lin2019nesterov} introduces the scale-invariant property to calculate the gradient over the input image scaled by a factor $1/2^i$ to generate transferable adversarial examples, where $i$ is a hyper-parameter. Holding the assumption of translation-invariant property, Translation-Invariant Method (TIM)~\cite{dong2019evading} adopts a set of translated images to calculate the gradient. Admix~\cite{wang2021admix} uses the original label of the input and calculates the gradient on the input image admixed with a small portion of each add-in image to improve the attack transferability. 
Different input transformation methods can be integrated with gradient-based attacks naturally. Our proposed \name can also be combined with these input transformations to further improve the attack transferability. 

\subsubsection{Model Ensemble Attacks}
Model ensemble attacks hold the hypothesis that if an adversary can fool multiple models, it will capture the intrinsic transferable information to fool the targeted model.
Liu~\etal~\cite{liu2017delving} propose to use an ensemble of multiple models, finding that the generated adversarial examples show better transferability. Dong~\etal~\cite{dong2018boosting} also use ensemble models by fusing their logit activations to further improve the transferability. 
Recently, apart from the above methods that attack a fixed set of models, multi-stage ensemble attacks have drawn attention~\cite{che2020new,hang2020ensemble,hu2022model}. Hang~\etal~\cite{hang2020ensemble} propose two types of model ensemble attacks, SCES and SPES. The former adopts a boosting structure while the latter employs the bagging structure. Their results show that the richer diversity of the ensemble model leads to better transferability. EnsembleFool~\cite{peng2021ensemblefool} dynamically selects the models according to their confidence output at the last iteration. SVRE~\cite{SVRE2022CVPR} is a recently proposed method that treats the iterative ensemble attack as a stochastic gradient descent optimization process and reduces the gradient variance of the ensemble models to take fully advantage of the ensemble attack.

\section{Methodology}
\label{sec:method}
This section first introduces four typical gradient-based attack methods, then illustrates our motivation 
and introduces the two key components of our method, followed by the detailed description of the proposed S-FGRM.

\subsection{Typical Gradient Optimization Attacks}
\label{sec:gradientMethods}
Gradient optimization methods are the mainstream of adversarial attacks, which also serve as the base for other categories of adversarial attacks. 

$\bullet$ Fast Gradient Sign Method (FGSM)~\cite{goodfellow2014explaining} is the basic single-step method for generating adversarial examples:
 \begin{equation}
      x^{adv}=x + \epsilon\cdot\text{sign}(\nabla_{x}J(x,y)),
 \end{equation}
 where $\nabla_{x}J(x,y)$ is the gradient of the loss function \wrt $x$.

$\bullet$ Iterative Fast Gradient Sign Method (I-FGSM)~\cite{kurakin2017adversarial} iteratively executes FGSM using a smaller step size: 
\begin{equation}
     \quad x_{t+1}^{adv} = x_t^{adv} + \alpha\cdot \text{sign}(\nabla_{x_t^{adv}}J(x_t^{adv},y^{true})),
\end{equation}
where $x_0^{adv}=x, \alpha=\epsilon/T$ is a small step size.

$\bullet$ Momentum Iterative Fast Gradient Sign Method (MI-FGSM)~\cite{dong2018boosting} introduces momentum into I-FGSM to boost the adversarial attacks:
 \begin{gather}
      g_{t+1}=\mu g_t + \frac{\nabla_{x_t^{adv}}J(x_t^{adv},y)}{||\nabla_{x_t^{adv}}J(x_t^{adv},y)||_1},\\
      x_{t+1}^{adv}=x_t^{adv}+\alpha\cdot \text{sign}(g_{t+1}),\notag
 \end{gather}
 where $g_0=0,x_0^{adv}=x$ and $\mu$ is a decay factor.

$\bullet$ Nesterov Iterative Fast Gradient Sign Method (NI-FGSM)~\cite{lin2019nesterov} integrates Nesterov's accelerated gradient into the iterative attack to look ahead and further improve the transferability:
 \begin{gather}
      x_t^{nes}=x_t^{adv}+\alpha\cdot\mu\cdot g_t,\notag\\
      g_{t+1} = \mu\cdot g_t + \frac{\nabla_{x_t^{nes}}J(x_t^{nes},y^{true})}{||\nabla_{x_t^{nes}}J(x_t^{nes},y^{true})||_1},\\
      x_{t+1}^{adv}=x_t^{adv} + \alpha\cdot \text{sign}(g_{t+1}).\notag
 \end{gather}


\subsection{Motivation of our Method}
\label{sec:motivation}
In the process of generating adversarial examples, gradient-based methods typically use the \emph{sign} function to estimate the gradient. Since the success of the first gradient-based method of FGSM~\cite{goodfellow2014explaining}, numerous studies have concentrated on attack strategies to generate more efficient and transferable adversarial examples based on FGSM, but take its basic \emph{sign} function for granted.

Though the \emph{sign} function guarantees an $l_\infty$ bound,  sufficient perturbations and a generally proper update direction, making it successful in various attacks, 
we observe that it also has some side effects, especially on the direction of the gradient update. The synthetic directions associated with each gradient value are limited since the output of the \emph{sign} function is either $\pm 1$ or 0, and 0 rarely happens for the real gradient values. Using \emph{sign} will result in an imprecise estimate on the gradient. As can be seen intuitively from Fig.~\ref{fig:sign} (a), the angle between \emph{sign}(\emph{g}) and \emph{black-box model} \emph{g'} is much larger than \emph{rescale}(\emph{g}). Based on the above analysis, we attempt to directly employ the distribution of the initial gradient by simply introducing the data rescaling method to the last step. It aims at retaining the difference between the gradient values as compared to the original \emph{sign} function. Our \textit{Fast Gradient Rescaling Method} is straightforward but effective, with little computational overhead analyzed below.   


One property of \emph{sign} is to produce enough perturbations. Without \emph{sign}, even tiny numerical inaccuracy might create 
big difference in the rescaling results due to the minimal values of the gradient. To compensate for this deficiency, we propose our \textit{Depth First Sampling Method}. By sampling around the input image, the negative influence of small value fluctuations will be alleviated. Unlike typical sampling methods, starting with the second sampling, we sample around the previous sampled image rather than the original input image. This strategy is similar to depth-first search. It not only mitigates the rescaling uncertainty but also investigates the model decision boundaries, which reduces overfitting to some extent 
and improves the attack transferability. To sum up, we propose the  \textit{Sampling-based Fast Gradient Rescaling Method (S-FGRM)} to further enhance the adversarial transferability by combining the above two strategies.

\subsection{Fast Gradient Rescaling Method}
\label{sec:gradient}
Typical gradient-based adversarial attack methods, like I-FGSM, MI-FGSM, NI-FGSM, \etc, are inextricably linked to the \emph{sign} function. There is no doubt that the \emph{sign} function provides sufficient perturbation for the gradient update, allowing crafted adversarial examples to 
approaching their targeted class more easily.  
Nevertheless, it is not the optimal or near-optimal solution. The synthetic directions produced by \emph{sign} are limited, as we have illustrated in Fig.~\ref{fig:sign} (b), which influences the speed of gradient convergence as well as the transferability. Gao~\etal~\cite{gao2021staircase} also observe the limitation of \emph{sign}, and they thereby sort all the gradient units and assign them with values designated for intervals. Their method sharply slows down the generation of adversarial examples due to the percentile calculation, which is not applicable in most complex scenarios. To develop a simple yet more efficient method, we introduce data rescaling into the gradient update and define the gradient rescaling function as follows:


\begin{gather}
\label{eq:rescaling}
    \operatorname{rescale}(g) = c * \operatorname{sign}(g)\odot f( \operatorname{norm}(\log_2\mid g\mid)),\\
    \operatorname{norm}(x) = \frac{x-\operatorname{mean}(x)}{\operatorname{std}(x)},\notag  ~~f(x)=\sigma=\frac{1}{1+e^{-x}}.\notag
\end{gather}

The notation $\mid g\mid$ means element-wise absolute value of gradient $g$. such design is to make the values within the scope of $log$ function. And it is also meaningful based on our observation that the distribution of original gradient is almost symmetric 
with 0 as the center. We utilize the logarithm of $\mid g\mid$ to base 2 to represent the fractional part of the gradient values in binary, thus the closely distributed values will be scattered. The original gradient is linearly transformed through the normalization. To further improve the transferability, data smoothing is performed at the end of the equation by mapping the intermediate result on $f(x)$. After several initial attempt, we find that the sigmoid function is efficient and effective for data smoothing within the scope between $[0,1]$ and $c$ is the rescale factor that controls the final scope between $[0,c]$. Since we operate on the absolute value and scatter them further, at last we will add their sign back.

Our rescale method leverages the value's difference and retain more accurate direction of the original gradients, \eg, for the actual 2D gradient $(0.8,10^{-8})$, the sign gradient will be $(1,1)$ while the resale gradient will be $(1.46,0.54)$ which still maintains the relative magnitude of the original gradients. $\log_2(x)$ is the key function that scatters the possible gradient values, e.g. ($\frac18, \frac14, \frac12$) will be scattered to ($-3$,$-2$,$-1$). And other outer wrapper functions work as norm or smooth function to make the scattered gradient smoother and more controllable, while maintaining monotonicity and producing consistent and enough perturbations.

Compared with Staircase~\cite{gao2021staircase}, our gradient rescaling method is more straightforward and more effective that do not have their expensive 
overhead on the sorting. The detailed analysis is shown in the next section. 


\subsection{Depth First Sampling Method}
\label{sec:depth}

\begin{figure}[t]
    \centering
    \includegraphics[width=0.6\linewidth]{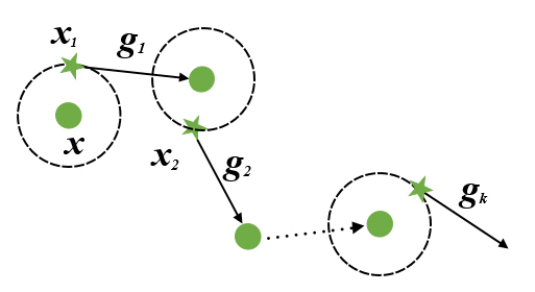}
    \caption{Illustration of the Depth First Sampling method. $x$ represents the input of the current iteration. Each sampling operation is performed on the previous sampled point. The average gradient of $g_{k}$ is accumulated to further boost the transferability. }
    \label{fig:dfs}
\end{figure}

In order to stabilize the effect of gradient rescaling, we adopt sampling to reduce the local fluctuation. Wu et al.~\cite{wu2020ACML} substitute each gradient with an averaged value during the iterations to alleviate the shattering on gradients. Inspired by the logic of depth-first search, we propose \textit{Depth First Sampling Method}. We define the potential samples around the current point in the input space as their neighbors. Instead of just sampling around the input images, we also take their neighbors into consideration. As illustrated in Fig.~\ref{fig:dfs}, when a sampling operation is completed, the next sampling center will shift to the point just sampled. Specifically, given a classifier $f$ with loss function $J$ and the input image $x \in \mathcal{X}$, our depth first sampling strategy is designed as follows:  
\begin{equation}
\label{eq:depth}
    g_t = \frac{1}{N+1}\sum_{i=0}^N\nabla J(x_t^{i},y;\theta),\\~~~\\
    x_t^{i+1} = x_t^{i}+\xi_i.
\end{equation}

Here $x_t^0=x$, $\xi_i \sim U[-(\beta\cdot\epsilon)^d,(\beta\cdot\epsilon)^d]$. $N$ is the sampling number. $\epsilon$ is the maximum perturbation in the current iteration and $\beta$ is a hyperparameter that determines the sampling range. By accumulating the gradient of sampled examples and benign image, the fluctuation effect is expected to be weakened and the transferability of the crafted examples will be enhanced. 

\begin{algorithm}[!t]
    \algnewcommand\algorithmicinput{\textbf{Input:}}
    \algnewcommand\Input{\item[\algorithmicinput]}
    \algnewcommand\algorithmicoutput{\textbf{Output:}}
    \algnewcommand\Output{\item[\algorithmicoutput]}
    \caption{SMI-FGRM}
    \label{alg:SMI-FGRM}
	\begin{algorithmic}[1]
		\Input A classifier $f$ with loss function $J$, parameters $\theta$; a benign example $x$ with ground-truth label $y$
		\Input Maximum perturbation $\epsilon$; number of iterations $T$
		\Input Number of sampling $N$
		\Output Adversarial example $x^{adv}$
		\State Initialize: $\alpha=\epsilon/T$, $x_0^{adv}=x$, $g_0=0$
		\For{$t = 0 \ \ {\rm to}\ \  T-1$}
		    \State Calculate the sampled gradient $\hat{g}_{t+1}$ by Eq. \ref{eq:depth}
		    \State Update $g_{t+1}$ by momentum
		    \State \qquad $g_{t+1}=\mu\cdot g_{t} + \frac{\hat{g}_{t+1}}{\left\|\hat{g}_{t+1}\right\|_1}$
		    \State Update $x_{t+1}^{adv}$ by the gradient rescaling method
		    \State \qquad $x_{t+1}^{adv}=x_{t}^{adv}+\alpha\cdot  \operatorname{rescale}(g_{t+1})$
		\EndFor
		\State $x^{adv}=x_T^{adv}$
	    \State \Return $x^{adv}$
	\end{algorithmic} 
\end{algorithm}

\subsection{The Final S-FGRM Algorithms}
\label{sec:sampled}
To gain a better grasp of our entire methodology, we incorporate the proposed method into MI-FGSM, denoted as Sampling-based Momentum Iterative Fast Gradient Rescaling Method (SMI-FGRM). Specific details are described in Algorithm~\ref{alg:SMI-FGRM}. 
Similarly, we could incorporate the proposed method into NI-FGSM, and obtain an enhanced method SNI-FGRM. In addition, 
almost any gradient-based attacks can integrate with our S-FGRM to gain more transferable attacks.

\section{Experiments}
\label{sec:exp}
This section starts with the experimental setup, then makes a comprehensive comparison with typical gradient-based attacks in a variety of attack settings, 
including the performance comparisons on attacking a single model, attacking with input transformation, attacking an ensemble of models,  attacking advanced defense models. 
Parameter and ablation studies are reported subsequently. 
In the end, we do further comparison with two attack methods, that are related to the two components in our method, respectively to show the superiority of our method.  

\begin{table*}[t]
  \centering
  \caption{Attack success rates (\%) of adversarial attacks on seven models in the single model setting. The adversarial examples are crafted on Inc-v3, Inc-v4, IncRes-v2, and Res-101 respectively. * indicates the white-box model.}
  \label{tab:S_FGRM}
  \resizebox{0.88\textwidth}{!}{
    \begin{tabular}{ l|l||c c c c c c c }
    \toprule
    Model & Attack & Inc-v3 & Inc-v4 & IncRes-v2 & Res-101 & Inc-v3$_{ens3}$ & Inc-v3$_{ens4}$ & IncRes-v2$_{ens}$ \\
    \midrule
    \multirow{4}{*}{Inc-v3} 
    & MI-FGSM & $\textbf{100.0}^{*}$ & 44.3 & 42.4 & 36.2 & 13.8 & 13.0 & 6.6\\
    & SMI-FGRM & $\textbf{100.0}^{*}$ & \textbf{82.0} & \textbf{81.1} & \textbf{73.6} & \textbf{44.8} & \textbf{41.3} & \textbf{23.1}\\
    \cmidrule{2-9}
    & NI-FGSM & $\textbf{100.0}^{*}$ & 51.3 & 49.9 & 40.6 & 12.8 & 12.9 & 6.6\\
    & SNI-FGRM & $\textbf{100.0}^{*}$ & \textbf{85.4} & \textbf{83.5} & \textbf{75.9} & \textbf{44.2} & \textbf{42.5} & \textbf{23.0}\\
    \midrule
    \multirow{4}{*}{Inc-v4} 
    & MI-FGSM & 55.3 & $\textbf{99.7}^{*}$ & 46.0 & 40.7 & 16.6 & 15.4 & 7.6\\
    & SMI-FGRM & \textbf{87.9} & $99.4^{*}$ & \textbf{83.6} & \textbf{76.5} & \textbf{55.5} & \textbf{52.2} & \textbf{35.6}\\
    \cmidrule{2-9}
    & NI-FGSM & 63.4 & $\textbf{100.0}^{*}$ & 52.4 & 45.0 & 15.5 & 14.3 & 6.9\\
    & SNI-FGRM & \textbf{89.5} & $99.9^*$ & \textbf{85.4} & \textbf{77.4} & \textbf{54.9} & \textbf{52.9} & \textbf{34.3}\\
    \midrule
    \multirow{4}{*}{IncRes-v2} 
    & MI-FGSM & 58.6 & 50.0 & $98.0^{*}$ & 46.0 & 21.2 & 16.1 & 10.7\\
    & SMI-FGRM & \textbf{86.0} & \textbf{84.5} & $\textbf{98.1}^{*}$ & \textbf{79.1} & \textbf{63.0} & \textbf{56.5} & \textbf{50.4}\\
    \cmidrule{2-9}
    & NI-FGSM & 64.1 & 54.1 & $\textbf{99.0}^{*}$ & 45.3 & 20.2 & 14.9 & 10.0\\
    & SNI-FGRM & \textbf{88.7} & \textbf{86.0} & $98.5^{*}$ & \textbf{81.8} & \textbf{64.6} & \textbf{55.7} & \textbf{50.8} \\
    \midrule
    \multirow{4}{*}{Res-101} 
    & MI-FGSM & 58.4 & 51.4 & 49.2 & $99.3^{*}$ & 23.8 & 21.3 & 12.8\\
    & SMI-FGRM & \textbf{85.5} & \textbf{81.3} & \textbf{82.2} & $\textbf{99.8}^{*}$ & \textbf{64.7} & \textbf{60.0} & \textbf{46.7} \\
    \cmidrule{2-9}
    & NI-FGSM & 65.5 & 59.8 & 56.0 & $99.4^{*}$ & 23.5 & 20.5 & 11.4\\
    & SNI-FGRM & \textbf{85.9} & \textbf{81.6} & \textbf{83.3} & $\textbf{99.8}^{*}$ & \textbf{64.4} & \textbf{59.9} & \textbf{48.0}\\
    \bottomrule
    \end{tabular}
    }

\end{table*}
\begin{table*}[ht]
  \centering
    \caption{Attack success rates (\%) of adversarial attacks on seven models in the single model setting with input transformations. The adversarial examples are crafted on Inc-v3, Inc-v4, IncRes-v2, and Res-101 respectively. * indicates the white-box model.}
    \label{tab:S_CT_FGRM}
    \resizebox{0.88\textwidth}{!}{
    \begin{tabular}{ l|l||c c c c c c c }
    \toprule
    Model & Attack & Inc-v3 & Inc-v4 & IncRes-v2 & Res-101 & Inc-v3$_{ens3}$ & Inc-v3$_{ens4}$ & IncRes-v2$_{ens}$ \\
    \midrule
    \multirow{8}{*}{Inc-v3} 
    & MI-CT-FGSM & $99.4^{*}$ & 84.0 & 81.5 & 78.1 & 68.2 & 64.2 & 46.2\\
    & SMI-CT-FGSM(\textbf{Ours}) & $\textbf{99.6}^{*}$ & 91.9 & 91.5 & \textbf{88.9} & 86.9 & 86.6 & 75.9\\
    & SMI-CT-FGRM(\textbf{Ours}) & $99.5^{*}$ & \textbf{93.3} & \textbf{92.8} & {88.4} & \textbf{88.9} & \textbf{86.8} & \textbf{77.8}\\
    \cmidrule{2-9}
    & NI-CT-FGSM & $99.3^{*}$ & 84.6 & 81.6 & 74.3 & 60.6 & 56.2 & 41.1\\
    & SNI-CT-FGSM(\textbf{Ours}) & $99.2^{*}$ & 93.4 & 91.6 & \textbf{89.5} & {86.7} & {86.1} & 77.2\\
    & SNI-CT-FGRM(\textbf{Ours}) & $\textbf{99.5}^{*}$ & \textbf{94.0} & \textbf{93.1} & {88.7} & \textbf{88.0} & \textbf{87.9} & \textbf{78.6}\\
    \midrule
    \multirow{8}{*}{Inc-v4} 
    & MI-CT-FGSM & 87.0 & $99.1^{*}$ & 84.1 & 78.6 & 70.4 & 69.1 & 58.0\\
    & SMI-CT-FGSM(\textbf{Ours}) & 91.9 & $98.7^{*}$ & 89.7 & 85.7 & 85.0 & {83.4} & {77.5}\\
    & SMI-CT-FGRM(\textbf{Ours}) & \textbf{93.8} & $\textbf{99.5}^{*}$ & \textbf{92.0} & \textbf{86.8} & \textbf{86.8} & \textbf{85.8} & \textbf{81.4}\\
    \cmidrule{2-9}
    & NI-CT-FGSM & 87.6 & $99.5^{*}$ & 81.9 & 77.0 & 66.7 & 61.5 & 49.6\\
    & SNI-CT-FGSM(\textbf{Ours}) & 93.1 & $99.6^{*}$ & 91.0 & 87.8 & 85.7 & 84.9 & {78.9}\\
    & SNI-CT-FGRM(\textbf{Ours}) & \textbf{94.9} & $\textbf{99.8}^{*}$ & \textbf{92.3} & \textbf{88.5} & \textbf{87.7} & \textbf{86.9} & \textbf{80.5}\\
    \midrule
    \multirow{8}{*}{IncRes-v2} 
    & MI-CT-FGSM & 88.1 & 86.7 & $97.2^{*}$ & 82.2 & 77.5 & 75.5 & 72.0\\
    & SMI-CT-FGSM(\textbf{Ours}) & 90.6 & 90.1 & $98.0^{*}$ & {87.0} & 87.1 & 86.5 & 85.2\\
    & SMI-CT-FGRM(\textbf{Ours}) & \textbf{93.9} & \textbf{92.5} & $\textbf{98.2}^{*}$ & \textbf{90.1} & \textbf{90.3} & \textbf{88.2} & \textbf{88.4}\\
    \cmidrule{2-9}
    & NI-CT-FGSM & 90.6 & 87.8 & $99.3^{*}$ & 83.4 & 74.1 & 68.3 & 64.4\\
    & SNI-CT-FGSM(\textbf{Ours}) & 93.2 & 92.4 & $99.4^{*}$ & 89.8 & 88.8 & 88.3 & 87.1\\
    & SNI-CT-FGRM(\textbf{Ours}) & \textbf{95.4} & \textbf{93.5} & $\textbf{99.6}^{*}$ & \textbf{91.7} & \textbf{92.0} & \textbf{89.7} & \textbf{89.5}\\
    \midrule
    \multirow{8}{*}{Res-101} 
    & MI-CT-FGSM & 85.7 & 82.0 & 82.4 & $98.5^{*}$ & 76.1 & 71.1 & 61.6\\
    & SMI-CT-FGSM(\textbf{Ours}) & 89.4 & 84.2 & 86.5 & $99.5^{*}$ & {89.1} & 86.2 & {81.6}\\
    & SMI-CT-FGRM(\textbf{Ours}) & \textbf{90.2} & \textbf{86.7} & \textbf{90.0} & $\textbf{99.7}^{*}$ & \textbf{90.2} & \textbf{87.9} & \textbf{84.0}\\
    \cmidrule{2-9}
    & NI-CT-FGSM & 87.3 & 83.2 & 85.6 & $99.1^{*}$ & 73.2 & 66.8 & 56.0\\
    & SNI-CT-FGSM(\textbf{Ours}) & \textbf{91.0} & 85.4 & 88.7 & $\textbf{99.7}^{*}$ & 88.8 & 87.2 & 81.5\\
    & SNI-CT-FGRM(\textbf{Ours}) & \textbf{91.0} & \textbf{86.2} & \textbf{89.0} & $\textbf{99.7}^{*}$ & \textbf{90.4} & \textbf{88.3} & \textbf{83.9}\\

    \bottomrule
    \end{tabular}
    }
\end{table*}

\subsection{Experimental Setup}

\textbf{Dataset}~
The dataset used for evaluation is from the ImageNet containing 1000 images randomly picked from the ILSVRC 2012 validation set~\cite{russakovsky2015imagenet}, which is  widely used in recent gradient-based attacks~\cite{dong2018boosting,lin2019nesterov,wang2021enhancing}.

\textbf{Models}~
We conduct experiments on seven 
popular models, including four normally trained models: Inception-v3 (Inc-v3)~\cite{szegedy2016rethinking}, Inception-v4 (Inc-v4), Inception-Resnet-v2 (IncRes-v2)~\cite{szegedy2017inception}, Resnet-v2-101 (Res-101)~\cite{he2016deep}, and three adversarially trained models: Inc-v3$_{ens3}$, Inc-v3$_{ens4}$ and IncRes-v2$_{ens}$~\cite{tramer2018ensemble}. 
We also 
include nine advanced defense models: HGD~\cite{liao2018defense}, R\&P~\cite{xie2018mitigating}, NIPS-r3, Bit-Red~\cite{xu2018feature}, JPEG~\cite{guo2018countering}, FD~\cite{liu2019feature}, ComDefend~\cite{jia2019comdefend}, RS~\cite{cohen2019certified} and NRP~\cite{naseer2020self}.

\textbf{Baselines}~
We select two popular gradient-based attack methods as our baselines, MI-FGSM~\cite{dong2018boosting} and NI-FGSM~\cite{lin2019nesterov}. 
We also integrate our methods with up-to-date input transformations, including DIM~\cite{xie2019improving}, TIM~\cite{dong2019evading}, SIM~\cite{lin2019nesterov} and their combination, CTM. 
Our boosted methods are denoted as SM(N)I-DI-FGRM, SM(N)I-TI-FGRM, SM(N)I-SI-FGRM, SM(N)I-CT-FGRM, respectively. 

\textbf{Hyperparameters}~
We follow the prior settings for hyper-parameters~\cite{dong2018boosting}. We set the maximum perturbation $\epsilon=16$ with the pixel values normalized to [0,1], the number of iterations $T=10$ and step size $\alpha=1.6$. 
We adopt the default decay factor $\mu=1.0$ for MI-FGSM and NI-FGSM. The transformation probability of DIM is set to 0.5. We adopt the Gaussian kernel with kernel size $7 \times 7$ for TIM. The number of scale copies is set to 5 for SIM. 
For our method, the parameters for sampling are set to $N=12$, $\beta=1.5$ and $c=2$.

\subsection{Attack a Single Model}
We first perform four adversarial attacks, including MI-FGSM, NI-FGSM, the proposed SMI-FGRM and SNI-FGRM on a single model to test the attack transferability. 
We generate adversarial examples on the first four normally trained networks respectively and evaluate them on the seven neural networks mentioned above. 
The attack success rates are presented in Table~\ref{tab:S_FGRM}. SMI-FGRM and SNI-FGRM maintain nearly 100$\%$ success rates in white-box setting. For black-box attacks, they surpass MI-FGSM and NI-FGSM by a large margin. For example, on the first and second rows, we craft adversarial examples on Inc-v3 model. Our proposed SMI-FGRM achieves $82.0\%$ success rate on Inc-v4 and $44.8\%$ success rate on Inc-v3$_{ens}$ while MI-FGSM only gets $44.3\%$ and $13.8\%$ corresponding success rates. The results demonstrate that S-FGRM boosts the two advanced gradient-based methods significantly and enhances the attack transferability.

\begin{table*}[tb]
  \centering
  \caption{Attack success rates (\%) on seven models in the ensemble-model setting. The adversarial examples are crafted on the ensemble model including Inc-v3, Inc-v4, IncRes-v2 and Res-101.}
  \label{tab:Ens}
  \resizebox{0.8\textwidth}{!}{
    \begin{tabular}{l||c c c c c c c }
    \toprule
    Attack & Inc-v3 & Inc-v4 & IncRes-v2 & Res-101 & Inc-v3$_{ens3}$ & Inc-v3$_{ens4}$ & IncRes-v2$_{ens}$ \\
    \midrule
    MI-FGSM & $100.0^{*}$ & $99.6^{*}$ & $99.5^{*}$ & $99.9^{*}$ & 47.9 & 43.1 & 27.9\\
    SMI-FGRM & ${99.9}^{*}$ & ${99.8}^{*}$ & ${99.9}^{*}$ & ${99.9}^{*}$ & \textbf{88.2} & \textbf{87.0} & \textbf{76.6}\\
    \cmidrule{1-8}
    NI-FGSM & ${100.0}^{*}$ & ${99.9}^{*}$ & $99.9^{*}$ & ${100.0}^{*}$ & 46.8 & 41.9 & 24.8\\
    SNI-FGRM & ${100.0}^{*}$ & ${99.9}^{*}$ & ${100.0}^{*}$ & $99.9^{*}$ & \textbf{89.5} & \textbf{88.4} & \textbf{77.6}\\
    \midrule 
    MI-CT-FGSM & $99.7^{*}$ & $98.8^{*}$ & $97.6^{*}$ & $99.9^{*}$ & 91.6 & 89.8 & 86.5\\
    SMI-CT-FGRM & ${99.9}^{*}$ & ${99.9}^{*}$ & ${99.7}^{*}$ & ${100.0}^{*}$ & \textbf{97.2} & \textbf{96.8} & \textbf{94.6}\\
    \cmidrule{1-8}
    NI-CT-FGSM & $99.8^{*}$ & ${99.9}^{*}$ & $99.3^{*}$ & ${99.9}^{*}$ & 92.3 & 89.7 & 84.4\\
    SNI-CT-FGRM & ${100.0}^{*}$ & ${99.9}^{*}$ & ${100.0}^{*}$ & ${99.9}^{*}$ & \textbf{98.5} & \textbf{97.4} & \textbf{95.7}\\
    \bottomrule
    \end{tabular}}

\end{table*}
\begin{table*}
  \centering
  \caption{Attack success rates (\%) of adversarial attacks on nine models with advanced defense mechanism.}
  \label{tab:Advanced}
  \resizebox{0.88\textwidth}{!}{
    \begin{tabular}{ l|l||c c c c c c c c c c }
    \toprule
    Model & Attack & HGD & R\&P & NIPS-r3 & Bit-Red & JPEG & FD & ComDefend & RS & NRP & Average \\
    \midrule
    \multirow{4}{*}{Inc-v3} 
    & MI-CT-FGSM  & 57.9 &  46.1 & 54.4 & 45.3 & 76.1 & 71.4 & 79.2 & 27.6 & 41.7 & 55.5 \\
    & SMI-CT-FGRM & \textbf{75.8} &  \textbf{70.6} & \textbf{78.4} & \textbf{66.9} & \textbf{88.2} & \textbf{84.9} & \textbf{90.3} & \textbf{44.2} & \textbf{67.5} & \textbf{74.0}\\
    \cmidrule{2-12}
    & NI-CT-FGSM  & 49.3 &  40.5 & 49.0 & 42.7 & 72.3 & 68.4 & 79.4 & 25.4 & 34.9 & 51.3\\
    & SNI-CT-FGRM & \textbf{76.9} &  \textbf{72.6} & \textbf{79.2} & \textbf{66.0} & \textbf{90.6} & \textbf{85.2} & \textbf{89.8} & \textbf{45.5} & \textbf{69.2} & \textbf{75.0}\\
    \midrule
    \multirow{4}{*}{Ens} 
    & MI-CT-FGSM  & 90.6 &  88.0 & 89.8 & 77.1 & 93.6 & 89.6 & 94.7 & 46.9 & 75.6 & 82.9\\
    & SMI-CT-FGRM & \textbf{95.1} &  \textbf{94.5} & \textbf{95.5} & \textbf{86.2} & \textbf{98.3} & \textbf{95.0} & \textbf{98.1} & \textbf{70.0} & \textbf{87.5} & \textbf{91.1}\\
    \cmidrule{2-12}
    & NI-CT-FGSM  & 90.8 & 85.3 & 89.1 & 72.0 & 96.1 & 89.0 & 95.4 & 43.5 & 70.3 & 81.3\\
    & SNI-CT-FGRM & \textbf{96.3} & \textbf{95.6} & \textbf{96.6} & \textbf{89.2} & \textbf{99.0} & \textbf{96.4} & \textbf{99.1} & \textbf{71.3} & \textbf{89.4} & \textbf{92.5} \\
    \bottomrule
    \end{tabular}
    }

\end{table*}

\subsection{Attack with Input Transformations}
Lin~\etal~\cite{lin2019nesterov} have demonstrated that the Composite Transformation Method (CTM), combined by the most 
advanced 
input transformations of DIM, TIM and SIM, improve the 
adversarial transferability significantly. 
Thus, we integrate our S-FGRM with CTM to further boost the transferability of gradient-based attacks. 
As depicted in Table~\ref{tab:S_CT_FGRM}, our SMI-CT-FGRM and SNI-CT-FGRM consistently outperform the baseline methods by $10\%\sim37\%$.
Here we also show the results of SMI-CT-FGSM and SNI-CT-FGSM that use FGSM with our DFS sampling. 
We can observe that adding the DFS samlping could help boost the transferbility, and replacing FGSM with FGRM could further improve the performance. 

\subsection{Attack an Ensemble of Models}
We continue to follow the ensemble attack settings~\cite{dong2018boosting}, and integrate our S-FGRM with the ensemble attack method to show that S-FGRM is capable of improving the transferability under multi-model setting. We craft adversarial examples on the ensemble of four normally trained models, \ie, Inc-v3, Inc-v4, IncRes-v2 and Res-101, with averaged logit outputs. As summarized in Table~\ref{tab:Ens}, our S-FGRM has better performance than the baseline attacks. It is worth noting that the proposed method 
 can improve the attack transferability of the baselines by more than $48\%$ for MI-FGSM and $52\%$ for NI-FGSM. In addition, when combined with CTM, S-FGRM boosts MI-CT-FGSM and NI-CT-FGSM on adversarially trained models by $6.9\%$ and $8.4\%$ on average, respectively. 

\subsection{Attack Advanced Defense Models}
To further validate the effectiveness, we evaluate our methods on nine models with advanced defenses, including the top-3 defense solutions in the NIPS competition: HGD (rank-1)~\cite{liao2018defense}, R\&P (rank-2)~\cite{xie2018mitigating}, NIPS-r3 (rank-3)) and six recently proposed defense models: Bit-Red~\cite{xu2018feature}, JPEG~\cite{guo2018countering}, FD~\cite{liu2019feature}, ComDefend~\cite{jia2019comdefend}, RS~\cite{cohen2019certified} and NRP~\cite{naseer2020self}. The results are organized in Table~\ref{tab:Advanced}. Notably, in the single model setting, our methods yield an average attack success rate of $74.0\%$ for SMI-CT-FGRM and $75.0\%$ for SNI-CT-FGRM on Inc-v3 model, outperforming the baselines for more than $18\%$ and $23\%$. In the multi-model setting, our methods achieve an average attack success rate of $91.1\%$ for SMI-CT-FGRM and $92.5\%$ for SNI-CT-FGRM. The consistent improvement shows great effectiveness and generalization of our S-FGRM for gradient-based attacks to achieve higher transferability.

\subsection{Parameter Study}
We conduct parameter study on the hyper-parameters of our method. To remove the influence of other factors, we tune $N$ and $\beta$ respectively and test the adversarial transferability on the Inc-v3 model without input transformations.

\subsubsection{On the number of sampled examples $N$}
\begin{figure}[t]
    \centering
    \begin{minipage}[c]{0.23\textwidth}
        \centering
        \includegraphics[width=0.85\linewidth]{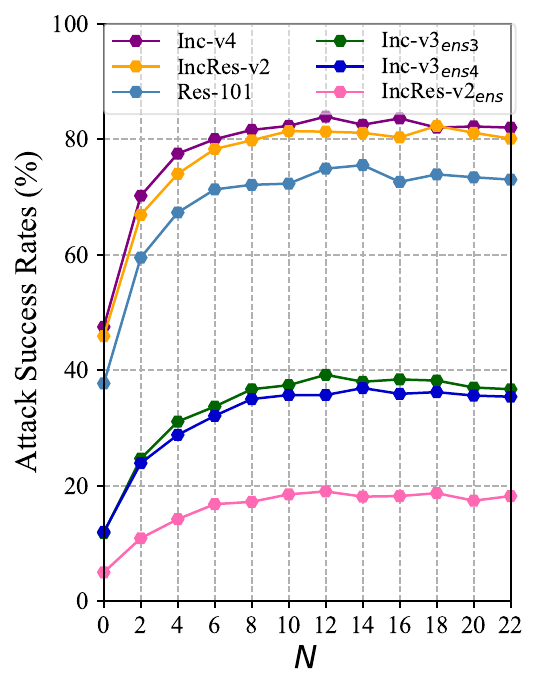}
        \label{fig:ablation_Nsmi}
    \end{minipage}
    \begin{minipage}[c]{0.23\textwidth}
        \centering
        \includegraphics[width=0.85\linewidth]{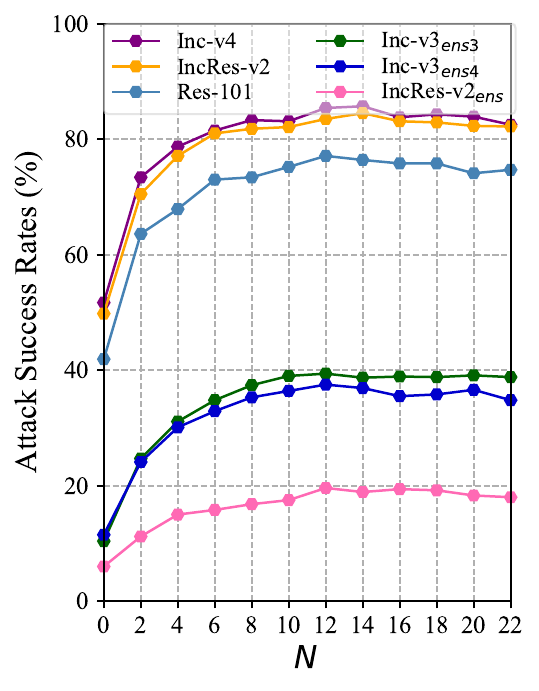}
        \label{fig:ablation_Nsni}
    \end{minipage}
    \caption{The attack success rates (\%) on the other six models with adversarial examples generated by SMI-FGRM (left) and SNI-FGRM (right) on Inc-v3 model for various sampling numbers. }
    \label{fig:ablation_N}
\end{figure}

We first analyze the effectiveness of the sampling number $N$ on the attack success rate of S-FGRM, as illustrated in Fig.~\ref{fig:ablation_N}. We integrate S-FGRM with MI-FGSM and NI-FGSM, respectively. 
The parameter of sampling range $\beta$ is fixed to 1.5, and we tune $N = 0,1,2,...,22$. 
When $N=0$, SMI-FGRM (SNI-FGRM) degrades to MI-FGRM (NI-FGRM).  
When the value of $N$ increases, the black-box attack success rate increases rapidly and reaches the peak at about $N=12$. We take $N = 12$ as the relatively optimal parameter in experiments since a bigger value of $N$ leads to higher computational overhead. Besides, we could also choose a smaller number of $N$ like 6, 8 or 10, the results are still significantly better than the baselines.

\begin{table*}[ht]
  \centering
    \caption{Attack success rates (\%) of adversarial attacks on seven models in the single model setting with different rescale methods. * indicates the white-box model.}
  \label{tab:FGRM}
    \resizebox{0.88\textwidth}{!}{
    \begin{tabular}{ l|c||c c c c c c c }
    \toprule
    Model & Attack & Inc-v3 & Inc-v4 & IncRes-v2 & Res-101 & Inc-v3$_{ens3}$ & Inc-v3$_{ens4}$ & IncRes-v2$_{ens}$ \\
    \midrule
    \multirow{4}{*}{Inc-v3} 
    & I-FGSM  & $\textbf{100.0}^{*}$ & 21.7 & 19.0 & 15.7 & 5.0 & 5.2 & 2.1\\
    & I-FGRM  & $\textbf{100.0}^{*}$ & 31.4 & 27.8 & 23.3 & 7.3 & 8.4 & 3.3\\
    & SI-FGSM & $\textbf{100.0}^{*}$ & 43.9 & 37.6 & 35.1 & 20.1 & 22.4 & 13.0\\
    & SI-FGRM & $\textbf{100.0}^{*}$ & \textbf{60.3} & \textbf{55.6} & \textbf{48.6} & \textbf{30.4} & \textbf{29.6} & \textbf{17.0}\\
    \midrule
    \multirow{4}{*}{Inc-v4} 
    & I-FGSM  & 31.2 & $\textbf{100.0}^{*}$ & 20.1 & 20.1 & 5.8 & 6.2 & 3.4\\
    & I-FGRM  & 40.7 & $99.8^{*}$ & 29.8 & 26.5 & 9.2 & 8.5 & 4.5\\
    & SI-FGSM  & 60.5 & $99.5^{*}$ & 46.1 & 41.3 & 29.9 & 26.5 & 20.0\\
    & SI-FGRM & \textbf{75.0} & $99.7^{*}$ & \textbf{61.6} & \textbf{56.4} & \textbf{39.8} & \textbf{37.4} & \textbf{27.1}\\
    \midrule
    \multirow{4}{*}{IncRes-v2} 
    & I-FGSM  & 32.5 & 24.7 & $98.2^{*}$ & 21.1 & 7.7 & 6.5 & 4.6\\
    & I-FGRM  &  42.3 & 34.9 & $98.5^{*}$ & 26.7 & 10.8 & 8.9 & 6.5\\
    & SI-FGSM  & 65.8 & 56.2 & $97.8^{*}$ & 51.1 & 35.7 & 31.7 & 28.9\\
    & SI-FGRM & \textbf{77.6} & \textbf{72.1} & $\textbf{98.6}^{*}$ & \textbf{62.9} & \textbf{49.9} & \textbf{43.6} & \textbf{40.6}\\
    \midrule
    \multirow{4}{*}{Res-101} 
    & I-FGSM  & 31.1 & 26.2 & 21.9 & $99.3^{*}$ & 7.9 & 8.3 & 5.3\\
    & I-FGRM  & 40.4 & 34.3 & 31.6 & $99.2^{*}$ & 14.4 & 13.1 & 7.7\\
    & SI-FGSM  & 55.7 & 45.4 & 47.9 & $\textbf{99.7}^{*}$ & 32.4 & 31.1 & 23.6\\
    & SI-FGRM & \textbf{66.7} & \textbf{58.5} & \textbf{60.1} & $\textbf{99.7}^{*}$ & \textbf{46.0} & \textbf{41.6} & \textbf{32.8}\\
    \bottomrule
    \end{tabular}}

\end{table*}


\begin{figure}[!ht]
    \centering
    \begin{minipage}[c]{0.23\textwidth}
        \centering
        \includegraphics[width=0.83\linewidth]{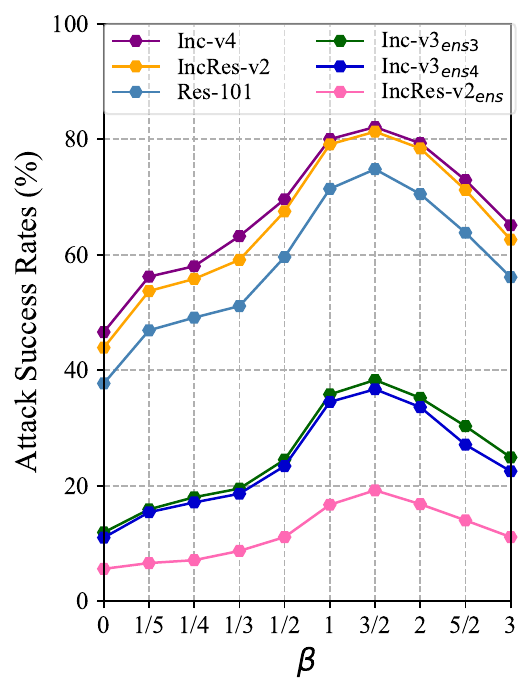}
        \label{fig:ablation_betasmi}
    \end{minipage}
    \begin{minipage}[c]{0.23\textwidth}
        \centering
        \includegraphics[width=0.83\linewidth]{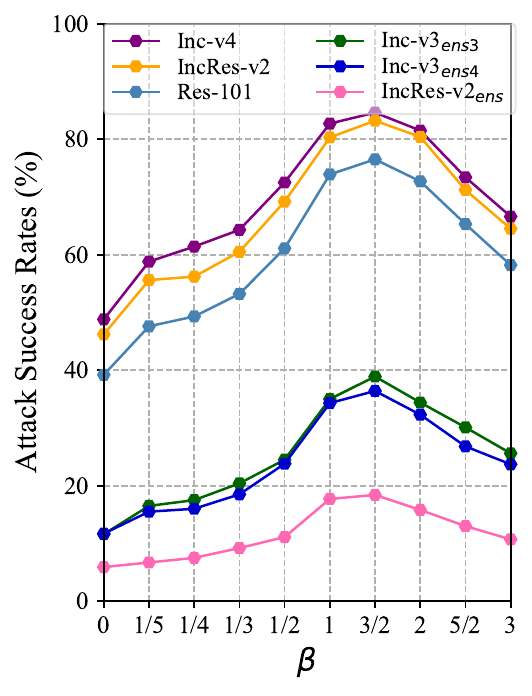}
        \label{fig:ablation_betasni}
    \end{minipage}
    \caption{The attack success rates (\%) on the other six models with adversarial examples generated by SMI-FGRM (left) and SNI-FGRM (right) on Inc-v3 model for various sampling ranges.}
    \label{fig:ablation_beta}
     \vspace{-1em}    
\end{figure}
\begin{figure}[t]
    \centering
    \begin{minipage}[c]{0.23\textwidth}
        \centering
        \includegraphics[width=0.85\linewidth]{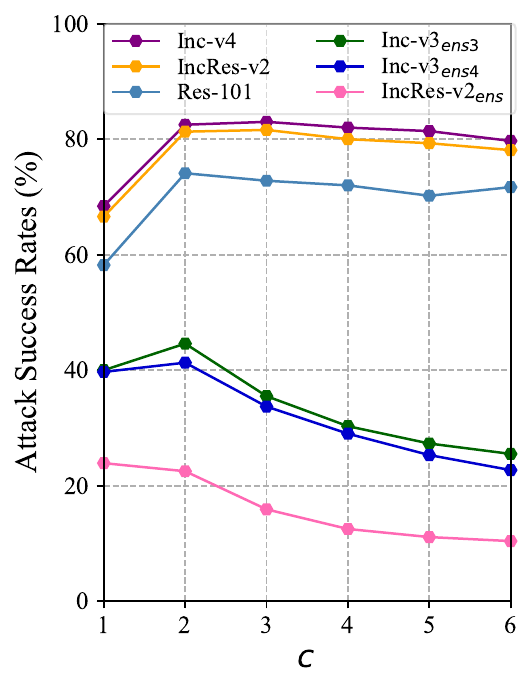}
        \label{fig:ablation_csmi}
    \end{minipage}
    \begin{minipage}[c]{0.23\textwidth}
        \centering
        \includegraphics[width=0.85\linewidth]{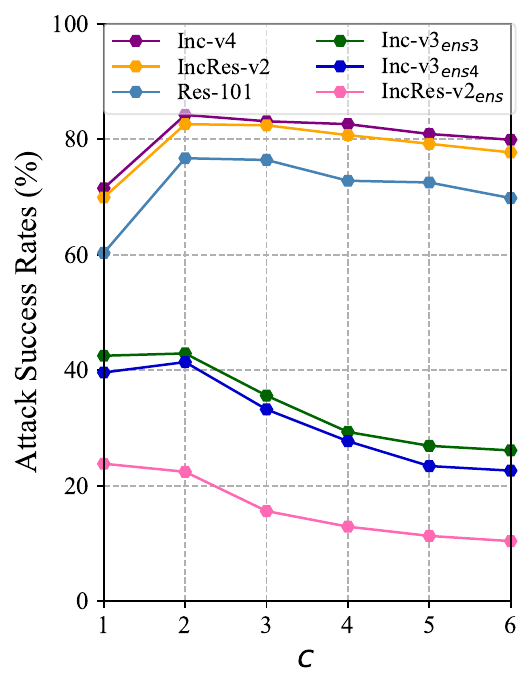}
        \label{fig:ablation_csni}
    \end{minipage}
    \caption{The attack success rates (\%) on the other six models with adversarial examples generated by SMI-FGRM (left) and SNI-FGRM (right) on Inc-v3 model for various rescale factors. }
    \label{fig:ablation_c}
\end{figure}

\subsubsection{On the size of sampling range $\beta$}
We then study the impact of the sampling range $\beta$. Similarly, we apply our S-FGRM to MI-FGSM and NI-FGSM, respectively, fix $N=12$, and let $\beta$ range from 0 to 3. $\beta=0$ denotes the corresponding FGRM. As shown in Fig.~\ref{fig:ablation_beta}, with the increment on the sampling range, the black-box success rate increases and remains stable after $\beta$ exceeds 1.5. When $\beta>1.5$, the performance decays gradually. 
Thus, we set the sampling range $\beta=1.5$ in experiments.

\subsubsection{On the size of rescale factor $c$}
We also test the influence of rescale factor $c$ in our \emph{rescale} function. It can be observed that 
the improvements of our S-FGRM are generally good for $c = 2$, as illustrated in Fig.~\ref{fig:ablation_c}. For adversarially trained models, the stability will decline if $c$ increases. We infer that a larger value of rescale factor may cause more pixels to be clipped, making it more difficult to generate effective perturbation. 
Hence, we choose $c = 2$ in our final experiments.

\subsection{Ablation Study}
We conduct ablation studies to validate the two components of our method, \ie, \textit{Fast Gradient Rescaling Method} and \textit{Depth First Sampling Method}. 
We test our method in the multi-step case to compare with I-FGSM, and the results are shown in Table~\ref{tab:FGRM}. 
We can observe that our I-FGRM exhibits clearly better results as compared with I-FGSM. 
After adding the depth first sampling method, our SI-FGRM gains further improvement on I-FGRM. The results show that our \emph{rescale} function exhibits better performance than the \emph{sign} function, and the DFS sampling further stabilizes the effects and boosts the performance. 

\begin{table*}[t]
  \centering
  \caption{Attack success rates (\%) of adversarial attacks with different sampling methods on seven models in the single model setting with input transformations. The adversarial examples are crafted on Inc-v3, Inc-v4, IncRes-v2, and Res-101 respectively. * indicates the white-box model.}
  \label{tab:sg_vs_dfsm}
  \resizebox{0.88\textwidth}{!}{
    \begin{tabular}{ l|l||c c c c c c c }
    \toprule
    Model & Attack & Inc-v3 & Inc-v4 & IncRes-v2 & Res-101 & Inc-v3$_{ens3}$ & Inc-v3$_{ens4}$ & IncRes-v2$_{ens}$ \\
    \midrule
    \multirow{4}{*}{Inc-v3} 
    & sgMI-CT-FGSM & $99.2^{*}$ & 86.9 & 86.6 & 83.4 & 86.3 & 85.6 & \textbf{76.5}\\
    & SMI-CT-FGSM & $\textbf{99.6}^{*}$ & \textbf{91.9} & \textbf{91.5} & \textbf{88.9} & \textbf{86.9} & \textbf{86.6} & 75.9\\
    \cmidrule{2-9}
    & sgNI-CT-FGSM & $\textbf{99.2}^{*}$ & 86.9 & 85.9 & 83.9 & 86.0 & 86.0 & 77.0\\
    & SNI-CT-FGSM & $\textbf{99.2}^{*}$ & \textbf{93.4} & \textbf{91.6} & \textbf{89.5} & \textbf{86.7} & \textbf{86.1} & \textbf{77.2}\\
    \midrule
    \multirow{4}{*}{Inc-v4} 
    & sgMI-CT-FGSM & 89.1 & $98.4^{*}$ & 84.0 & 81.4 & 84.8 & \textbf{83.8} & 77.2\\
    & SMI-CT-FGSM & \textbf{91.9} & $\textbf{98.7}^{*}$ & \textbf{89.7} & \textbf{85.7} & \textbf{85.0} & 83.4 & \textbf{77.5}\\
    \cmidrule{2-9}
    & sgNI-CT-FGSM & 87.7 & ${98.3}^{*}$ & 84.1 & 81.1 & 85.0 & 83.7 & 77.7\\
    & SNI-CT-FGSM & \textbf{93.1} & $\textbf{99.6}^{*}$ & \textbf{91.0} & \textbf{87.8} & \textbf{85.7} & \textbf{84.9} & \textbf{78.9}\\
    \midrule
    \multirow{4}{*}{IncRes-v2} 
    & sgMI-CT-FGSM & 87.1 & 84.8 & $97.0^{*}$ & 84.6 & \textbf{87.6} & \textbf{87.0} & 84.4\\
    & SMI-CT-FGSM & \textbf{90.6} & \textbf{90.1} & $\textbf{98.0}^{*}$ & \textbf{87.0} & 87.1 & 86.5 & \textbf{85.2}\\
    \cmidrule{2-9}
    & sgNI-CT-FGSM & 88.0 & 86.3 & ${97.5}^{*}$ & 85.0 & 88.1 & 86.2 & 85.7\\
    & SNI-CT-FGSM & \textbf{93.2} & \textbf{92.4} & $\textbf{99.4}^{*}$ & \textbf{89.8} & \textbf{88.8} & \textbf{88.3} & \textbf{87.1} \\
    \midrule
    \multirow{4}{*}{Res-101} 
    & sgMI-CT-FGSM & 81.9 & 73.0 & 78.5 & $99.1^{*}$ & 86.5 & 84.1 & 78.6\\
    & SMI-CT-FGSM & \textbf{89.4} & \textbf{84.2} & \textbf{86.5} & $\textbf{99.5}^{*}$ & \textbf{89.1} & \textbf{86.2} & \textbf{81.6} \\
    \cmidrule{2-9}
    & sgNI-CT-FGSM & 81.5 & 71.5 & 78.2 & $98.7^{*}$ & 85.2 & 83.7 & 77.2\\
    & SNI-CT-FGSM & \textbf{91.0} & \textbf{85.4} & \textbf{88.7} & $\textbf{99.7}^{*}$ & \textbf{88.8} & \textbf{87.2} & \textbf{81.5}\\
    \bottomrule
    \end{tabular}
    }

\end{table*}
\subsection{Further Comparison}
We further compare the two components of S-FGRM with SGA (Smooth Gradient Attack)~\cite{wu2020ACML} and Staircase~\cite{gao2021staircase} respectively to show the superiority of our method.

\subsubsection{DFSM vs. SGA}
SGA samples around the image by adding Gaussian noise to the input
and substitutes each gradient with an averaged value. We set the sampling number $N=20$ to gain SGA's best performance and set $N=12$ for our method. As depicted in Table~\ref{tab:sg_vs_dfsm}, our SMI-CT-FGSM consistently outperforms sgMI-CT-FGSM as the same for SNI-CT-FGSM over sgNI-CT-FGSM. DFSM can take fewer samples and achieve better better adversarial transferability. The great performance stated above convincingly validates the effectiveness of proposed DFSM.

\subsubsection{FGRM vs. Staircase}
To further investigate the effectiveness of our \emph{rescale} function, we compare with the staircase sign method, termed S$^2$M, which is a recent work that also focuses on the improvement of the \emph{sign} function. 
This method heuristically divides the gradient sign into several segments according to the sorted values of the gradient units, and then assigns each segment with a staircase weight for better crafting adversarial perturbation. Specifically, it assigns the staircase weights according to the $p$-th percentile of the gradient units at each iteration where $p$ depends on the number $K$ of staircases.
We set $K=64$ to gain Staircase's best performance and set the sampling number $N=12$ for our method. As shown in the results, our FGRM can achieve comparable performance to Staircase. Note that our method is more efficient than Staircase because we do not need to sort the gradient values to calculate the percentile. Considering the batch input whose size is $S = B\times H\times W\times C$, where $B$ is the batch size and $H$, $W$, $C$ represent the height, width and channel of an image, the Staircase method needs $O(S\log S)$ computational cost per iteration while our rescale method only needs $O(S)$. 


\begin{table*}[ht]
  \caption{Attack success rates (\%) of adversarial attacks with different perturbation generation functions on seven models in the single model setting with input transformation. The adversarial examples are crafted on Inc-v3, Inc-v4, IncRes-v2, and Res-101 respectively. * indicates the white-box model.}
  \label{tab:C_MI_CT}
  \centering
    \resizebox{0.88\textwidth}{!}{
    \begin{tabular}{ l|l||c c c c c c c }
    \toprule
    Model & Attack & Inc-v3 & Inc-v4 & IncRes-v2 & Res-101 & Inc-v3$_{ens3}$ & Inc-v3$_{ens4}$ & IncRes-v2$_{ens}$ \\
    \midrule
    \multirow{6}{*}{Inc-v3} 
    & MI-CT-FGSM &98.0*&83.3&80.3&75.0&64.7&61.9&45.1\\
    & MI-CT-FGS$^2$M &99.4*&85.6&81.5&77.1&\textbf{70.7}&64.1&\textbf{48.9}\\
    \cmidrule{2-9}
    & MI-CT-FGRM(2) &99.3*&86.2&82.7&77.8&67.5&\textbf{64.9}&48.7\\
    & MI-CT-FGRM(3) &99.2*&86.2&83.0&\textbf{78.6}&62.6&58.4&41.0\\
    & MI-CT-FGRM(4) &99.5*&86.3&83.6&78.3&58.9&54.5&38.7\\
    & MI-CT-FGRM(5) &99.1*&\textbf{87.2}&\textbf{83.8}&78.1&54.7&50.0&34.6\\
    \midrule
    \multirow{6}{*}{Inc-v4} 
    & MI-CT-FGSM &85.1&98.9*&83.0&74.4&69.6&66.1&56.0\\
    & MI-CT-FGS$^2$M &87.7&99.3*&84.5&78.1&\textbf{73.4}&\textbf{69.4}&\textbf{59.4}\\
    \cmidrule{2-9}
    & MI-CT-FGRM(2) &88.5&98.9*&84.0&78.1&71.9&69.3&58.2\\
    & MI-CT-FGRM(3) &88.6&99.4*&85.2&\textbf{79.7}&67.9&65.7&54.2\\
    & MI-CT-FGRM(4) &\textbf{89.1}&99.1*&\textbf{85.3}&79.1&65.8&61.2&48.9\\
    & MI-CT-FGRM(5) &88.9&99.4*&84.0&79.6&62.4&58.8&45.2\\
    \midrule
    \multirow{6}{*}{IncRes-v2} 
    & MI-CT-FGSM &88.2&85.3&96.8*&82.2&76.7&72.4&69.5\\
    & MI-CT-FGS$^2$M &90.5&88.5&98.3*&84.2&\textbf{81.4}&\textbf{78.2}&\textbf{75.7}\\
    \cmidrule{2-9}
    & MI-CT-FGRM(2) &89.9&88.3&97.9*&83.9&\textbf{81.4}&76.6&\textbf{75.7}\\
    & MI-CT-FGRM(3) &90.7&88.6&98.8*&85.8&79.9&75.5&69.7\\
    & MI-CT-FGRM(4) &91.4&\textbf{90.8}&98.7*&\textbf{86.5}&76.5&69.6&63.1\\
    & MI-CT-FGRM(5) &\textbf{92.5}&\textbf{90.8}&99.1*&86.1&74.7&67.9&59.2\\
    \midrule
    \multirow{5}{*}{Res-101} 
    & MI-CT-FGSM &84.2&80.9&81.3&99.2*&74.5&70.1&60.7\\
    & MI-CT-FGS$^2$M &86.0&82.6&85.6&99.2*&76.5&\textbf{73.1}&\textbf{65.2}\\
    \cmidrule{2-9}
    & MI-CT-FGRM(2) &86.8&82.2&84.7&98.7*&\textbf{77.4}&72.4&63.0\\
    & MI-CT-FGRM(3) &88.1&84.0&84.6&98.8*&74.7&68.8&59.5\\
    & MI-CT-FGRM(4) &\textbf{88.2}&\textbf{85.5}&85.3&99.1*&71.3&66.4&55.0\\
    & MI-CT-FGRM(5) &87.8&84.3&\textbf{86.1}&99.1*&70.2&64.0&49.8\\
    \bottomrule
    \end{tabular}
    }

\end{table*}


\section{Conclusion}
\label{sec:conclusion}
In this work, we revisited the basic operation of the \emph{sign} function in gradient-based adversarial attack methods, and discussed its limitation caused by the inaccurate approximation of the true gradients. Based on our observation, we proposed a simple yet more efficient method called the Sampling-based Fast Gradient Rescaling Method (S-FGRM) to enhance the adversarial transferability for black-box attacks. Specifically, we introduced a data rescaling method to the gradient update and removed the local fluctuation by the depth first sampling method. 
In this way, S-FGRM can further improve the adversarial transferability. We then incorporated our S-FGRM method into typical gradient-based attacks. Our methods surpassed the 
advanced attack methods, MI-FGSM and NI-FGSM, by a large margin. 
Extensive experiments validated the efficacy and broad applicability of our method. 
S-FGRM is generally applicable to any gradient-based attack and can be combined with other input transformation or ensemble attacks to enhance the adversarial transferability, which deserves more exploration in future work.




\bibliographystyle{IEEEtran}
\bibliography{main}


\end{document}